\documentclass[letterpaper, 10 pt, conference]{IEEEtran}  %

\usepackage[letterpaper,left=.75in,right=.75in,top=.75in,bottom=.78in]{geometry}

\usepackage{amsmath} %
\usepackage{amssymb}  %
\usepackage[pdftex]{graphicx}
\graphicspath{{figures/}}
\usepackage{hyperref}
\hypersetup{
    colorlinks=true,
    linkcolor=black,
    citecolor=black,
    filecolor=black,
    urlcolor=black,
}
\usepackage[T1]{fontenc}
\usepackage[utf8]{inputenc}
\usepackage[english]{babel}
\usepackage[export]{adjustbox}
\usepackage[font=small]{caption}
\usepackage{subcaption}
\usepackage[colorinlistoftodos]{todonotes}
\usepackage{lipsum}

\usepackage{placeins}%

\newcommand{\argmin}{\operatornamewithlimits{argmin}}

\title{\vspace{.25in} \LARGE \bf
Generating Comfortable, Safe and Comprehensible Trajectories for Automated Vehicles in Mixed Traffic
}

\author{%
\IEEEauthorblockN{Maximilian Naumann\IEEEauthorrefmark{1}\IEEEauthorrefmark{2}, 
Martin Lauer\IEEEauthorrefmark{2} and 
Christoph Stiller\IEEEauthorrefmark{1}\IEEEauthorrefmark{2}}%
\IEEEauthorblockA{\IEEEauthorrefmark{1}FZI Research Center for Information Technology,
Mobile Perception Systems,
Karlsruhe, Germany\\
{\tt\small naumann@fzi.de}}%
\IEEEauthorblockA{\IEEEauthorrefmark{2}Karlsruhe Institute of Technology (KIT),
Institute of Measurement and Control,
Karlsruhe, Germany\\
{\tt\small \{martin.lauer,stiller\}@kit.edu}}}

\usepackage{url}
\usepackage[absolute,showboxes]{textpos}

\TPMargin{5pt}

\newcommand{\copyrightstatement}{
    \begin{textblock}{0.84}(0.08,0.02)    %
         \noindent
         \footnotesize
         \copyright  2018 IEEE.
         Personal use of this material is permitted. Permission from IEEE must be obtained for all other uses, in any current or future media, including reprinting/republishing this material for advertising or promotional purposes,creating new collective works, for resale or redistribution to servers or lists, or reuse of any copyrighted component of this work in other works. 

         \vspace{2mm}
         \noindent
         Appeared in:  Proc. IEEE Intl. Conf. Intelligent Transportation Systems (ITSC), pp. 575--582, Maui, Hawaii, USA, Nov 2018.
         
         \noindent
         DOI: \url{https://doi.org/10.1109/ITSC.2018.8569658}
    \end{textblock}
}

\begin{document}

\copyrightstatement

\maketitle
\thispagestyle{empty}
\pagestyle{empty}

\begin{abstract}
While motion planning approaches for automated driving often focus on safety and mathematical optimality with respect to technical parameters,
they barely consider convenience, perceived safety for the passenger and comprehensibility for other traffic participants.
For automated driving in mixed traffic, however, this is key to reach public acceptance.
In this paper, we revise the problem statement of motion planning in mixed traffic:
Instead of largely simplifying the motion planning problem to a convex optimization problem,
we keep a more complex probabilistic multi agent model and strive for a near optimal solution.
We assume cooperation of other traffic participants, yet being aware of violations of this assumption.
This approach yields solutions that are provably safe in all situations, and convenient and comprehensible in situations that are also unambiguous for humans. 
Thus, it outperforms existing approaches in mixed traffic scenarios, as we show in simulation.
\end{abstract}
\section{Introduction}
\label{sec:introduction}

In the last decades, tremendous progress has been achieved in the field of automated driving \cite{BenglerDietmayerFarberMaurerStillerWinner2014ITSM}.
Recently, some companies even announced the large-scale deployment of hundreds of fully automated vehicles in public road traffic, i.e. vehicles that fulfill SAE level 5 \cite{SAE}.

Obviously, vehicles must operate safely under all encountered circumstances to be accepted by the public.
Shalev-Shwartz et al. motivate that the fatality rate should be reduced by three orders
of magnitude, i.e. from $10^{-6}/h$ to $10^{-9}/h$ \cite{mobileye_whitepaper_2017}.
However, it is also obvious that this safety improvement must not result in a vast lack of comfort and utility. 
One can assume, that people will only use automated cars, if they improve safety without loosing too much comfort or utility.

\begin{figure}
	\centering
	\begin{subfigure}[t]{0.9\linewidth}
	\includegraphics[trim={0 0.8cm 0 0},clip,width=\linewidth]{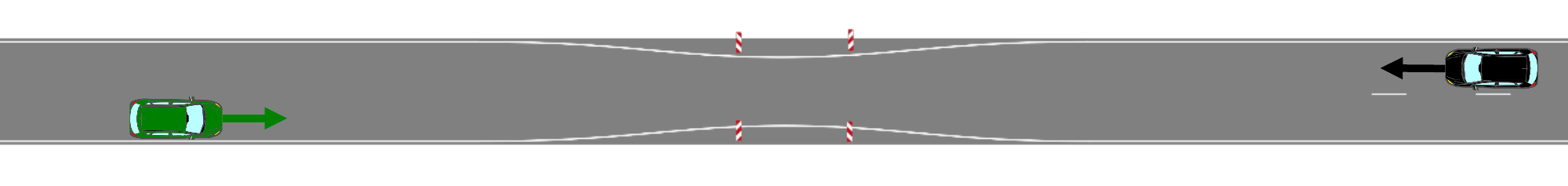}
	\caption{Scenario: Narrowing}
	\label{fig:page1fig_a}
	\end{subfigure}
	\begin{subfigure}[t]{0.49\linewidth}
	\includegraphics[width=\linewidth]{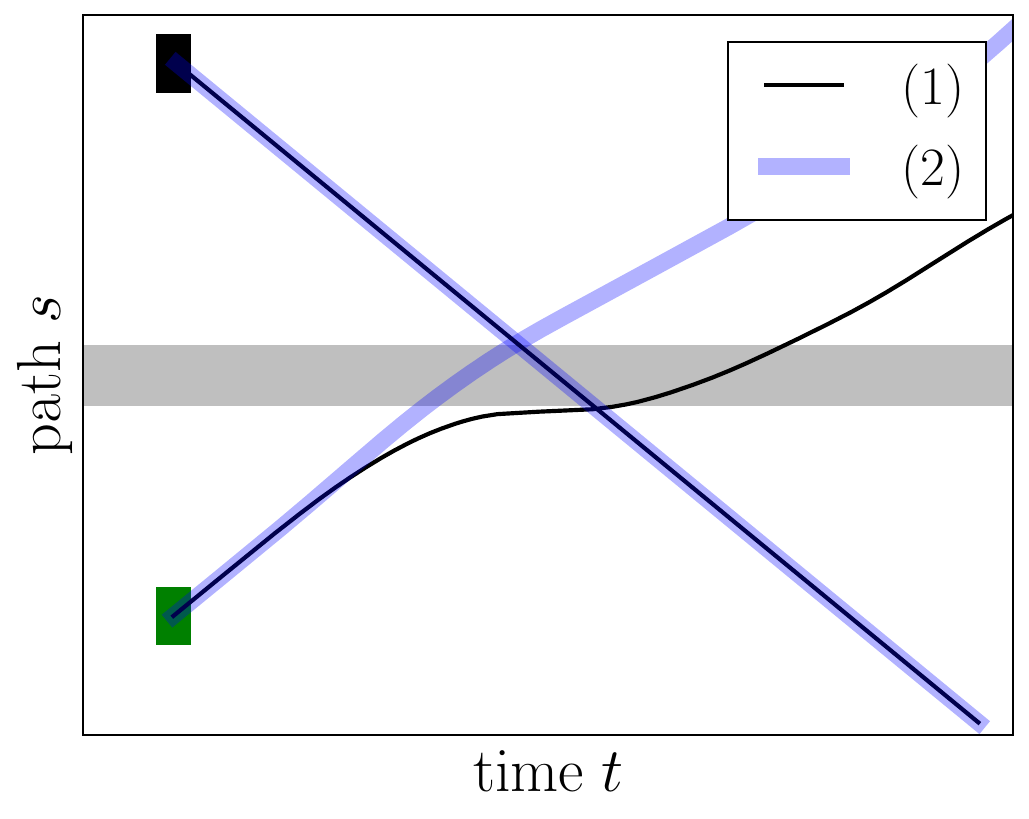}
	\caption{Classical Approach}
	\label{fig:page1fig_b}
	\end{subfigure}
	\begin{subfigure}[t]{0.49\linewidth}
	\includegraphics[width=\linewidth]{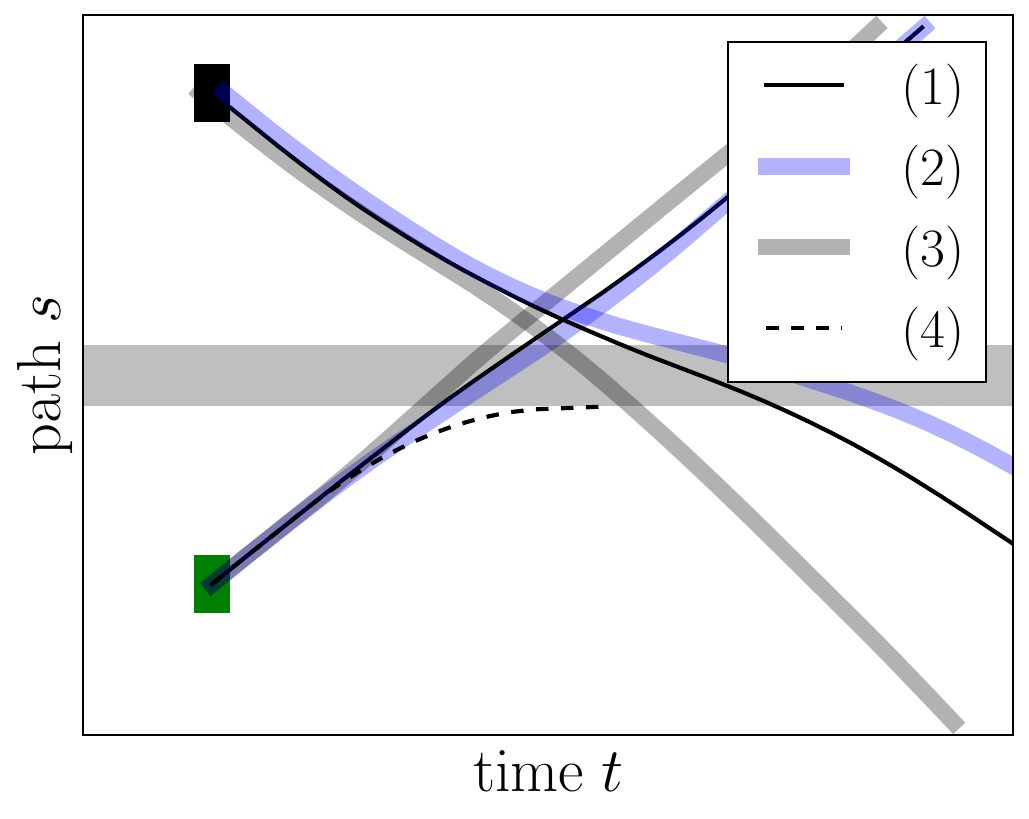} 
	\caption{Cooperative Approach}
	\label{fig:page1fig_c}
	\end{subfigure}
	
	\caption{Time-path-diagrams of the plan of the ego vehicle (green). (b) Classical approaches treat other traffic participants as obstacles: Driving second (1) through the potential collision zone (grey) appears to be the solution with the least costs.
	(c) The proposed cooperative approach treats other traffic participants as cooperative agents: Driving first (1) is globally optimal, also for different driver types (2), (3) and we have a comfortable fallback option (4).}
	\label{fig:page1fig}
\end{figure}

In order to operate safely, an automated vehicle undoubtably needs a comprehensive, redundant perception.
Further, it needs a motion planning module that is able to guarantee safety under some constraints concerning the perception accuracy and the behavior of other traffic participants.
In order to also operate comfortably, the vehicle needs to analyze the situation and predict its future evolution.
The latter is often provided by a module called \textit{situation prediction}, which predicts the motion of other traffic participants and provides it to the motion planning module.
In the motion planning, others are then treated as obstacles that need to be avoided.
This hierarchical design and the treatment as obstacles, however, does not account for the fact that the motion of the ego vehicle potentially affects the motion of other traffic participants (cf. Fig. \ref{fig:page1fig}).

In this work, we thoroughly review the problem statement of behavior generation and motion planning for automated vehicles.
We pay special attention to the inevitable uncertainties that motion planning approaches have to deal with: the uncertainty in the perception and in the prediction of other traffic participants.
With these considerations, as an extension to our previous work \cite{naumann2017cooperativePlanning}, we propose an integrated approach to generate comfortable, safe and comprehensible trajectories for automated vehicles in mixed traffic:
We explicitly model the strategy of other traffic participants and check their model compliance.
Thereby we facilitate cooperation with human driven vehicles in many situations and thus largely increase driving comfort and convenience.
In parallel, we run a safety planner, ensuring that we never cause a collision.
The remainder of this paper is structured as follows:
The problem statement and related work concerning optimization, uncertainty consideration and safety guarantees are presented in the following sections.
Subsequently, our approach is presented in Section \ref{sec:optimization_problem}.
It is evaluated in Section \ref{sec:results_and_evaluation}, using our simulation framework \textit{CoInCar-Sim} \cite{Naumann2018Sim}.

\section{Problem Statement}
\label{sec:basic_idea_assumptions}

Presumably, the goal of every traffic participant is to travel to a certain destination in a convenient way.
Having this in mind, we introduce our notation and motivate the problem formulation. 

\subsection{Challenges}
The task of finding an optimal plan from a start state to a desired goal state is called \textit{motion planning}.
The common approach to motion planning is the formulation of an optimization problem:
A certain cost functional for state transitions is defined, along with some constraints.
The plan with the lowest cost that does not violate any constraint is determined by the minimization of the cost functional.
Analogously, a reward functional can be maximized.

Considering automated driving, the motion planning task is challenging due to several circumstances:
\begin{enumerate}
\item \textit{cost formulation}: comfort and perceived safety is difficult to express numerically
\item \textit{uncertain perception}: the perception of the status quo is subject to uncertainties
\item \textit{uncertain prediction}: the evolution of the scene is unknown and it depends on other agents' and the own plan
\end{enumerate}

\subsection{Notation}
In the remainder of this paper, we employ the following notation: %

Considering the time $t$, we refer to
\begin{itemize}
\item $t_0$  as the start time of the decision problem
\item $\Delta t_\mathrm{action}$  as the time that actions (introduced below) are carried out in
\item $\Delta t_\mathrm{plan}$  as the planning horizon
\item $t_i = t_0 + i \Delta t_\mathrm{action}$ as the (re)planning points in time
\item $N_{\mathrm{a, pl}} = \frac{\Delta t_\mathrm{plan}}{\Delta t_\mathrm{action}}$ as the number of actions within the planning horizon.
\end{itemize}

Considering the decision problem itself, we refer to 

\begin{itemize}
\item $\mathcal{S}$ as the set of possible world states, with the state vector $\mathbf{s} \in \mathcal{S}$, containing a finite number of state variables $s$ describing the current scene
\item $\mathcal{A}$ as the set of possible actions, with the action vector $\mathbf{a} \in \mathcal{A}$, containing a finite number of actions $a$ that are performed and that influence the evolution of the state vector over time, from $\mathbf{s}_{t_i}$ to $\mathbf{s}_{t_{i+1}}$
\item $P_\mathbf{a}(\mathbf{s}, \mathbf{s'}) = \Pr(\mathbf{s}_{t_{i+1}} =  \mathbf{s'} | \mathbf{s}_{t_i} = \mathbf{s}, \mathbf{a}_{t_i} = \mathbf{a} )$ as the transition probability, describing the probability that actions $\mathbf{a}$ at time $t_i$ in state $\mathbf{s}$ will lead to state $\mathbf{s'}$ at time $t_{i+1}$. In case of deterministic transitions, the probability density function $P_\mathbf{a}(\mathbf{s}, \mathbf{s'})$ turns into an ordinary function $f$ with $\mathbf{s'} = f(\mathbf{s},\mathbf{a})$. Either formulation relies on the Markov assumption, that the effects of an action $\mathbf{a}$ taken in a state $\mathbf{s}$ depend only on that state and not on its prior history. 
\item $g(\mathbf{s}_{t_i}, \mathbf{s}_{t_{i+1}}, \mathbf{a})$ as the cost function, describing the cost of a state transition from $\mathbf{s}_{t_i}$ to $\mathbf{s}_{t_{i+1}}$ taking action $\mathbf{a}$
\end{itemize}

\subsection{Motivating the MMDP Perspective}

In order to consider the problem as a classical \textit{Markov Decision Process (MDP)}, one must be able to solely decide which action $\mathbf{a}$ is taken. 
Actions performed by other traffic participants would not be modeled explicitly. 
In the following, we distinguish between scalar $a$ for the action of one vehicle and $\mathbf{a}$ for the actions of multiple vehicles.
In case of a classical MDP, the actions of other traffic participants
would be covered in the transition probability $P_{a}(\mathbf{s}, \mathbf{s'})$.
The optimal policy $a = \pi(\mathbf{s})$ can then be calculated for example via minimizing 
\begin{itemize}
	\item the expected \textit{finite horizon} costs
	\begin{equation}\label{eqn:fin_hor_cost}
	\mathbf{E}\left(
		\sum_{i = 0}^{N_{\mathrm{a, pl}}} 
		g(\mathbf{s}_{t_i}, \mathbf{s}_{t_{i+1}}, a_{t_i})
	\right)
	\end{equation}
	where \textit{final costs} can be added to account for the time not covered by the horizon,
	\item or the expected \textit{cumulative discounted} costs 
	\begin{equation}\label{eqn:cumulative_cost}
	\mathbf{E}\left(
		\sum_{i = 0}^{\infty}
		\gamma^{i} g(\mathbf{s}_{t_i}, \mathbf{s}_{t_{i+1}}, a_{t_i})
	\right)
	\end{equation}
	with discount factor $\gamma \in \lbrack 0,1 ) $.
\end{itemize}

However, in road traffic, every traffic participant decides over its future state by its own action\footnote{Assuming collision free motion of every traffic participant.}.
If we know the current state $\mathbf{s}_{t_i}$ including all objects and the environmental condition, and the actions of all objects $\mathbf{a}_{t_i}$, the transition to $\mathbf{s}_{t_{i+1}}$ can be considered deterministic, except for the environmental condition.
The change of environmental conditions can be either neglected or considered as known throughout the decision problem, as the traffic participants do not influence it.

Thus, we model this problem as a \textit{Multi Agent Markov Decision Process (MMDP)} with deterministic transitions:
We choose a subset of the state vector, that defines the state of the ego vehicle $\mathbf{s}^\mathrm{ego}$, one that defines the state of other traffic participants $\mathbf{s}^\mathrm{other} = (\mathbf{s}^\mathrm{obj1}, \mathbf{s}^\mathrm{obj2}, ...)$ and one that describes the environment $\mathbf{s}^\mathrm{env}$.
Further, the action vector contains one action per traffic participant $\mathbf{a} = (a^\mathrm{ego}, a^\mathrm{obj1}, a^\mathrm{obj2}, ...)$.
However, we can solely decide the action of the ego vehicle.

The sequence of actions $a^\mathrm{obj}$ of an object with initial state $\mathbf{s}^\mathrm{obj}_{t_0}$ can then be described by its trajectory 
$\mathbf{x}^\mathrm{obj}(t) = (x(t), y(t))^T, \; t \in [t_0, t_0+\Delta t_\mathrm{plan}]$, which is a mapping $\mathbb{R} \rightarrow \mathbb{R}^2$ and a flat output for a kinematic vehicle model, %
along with its time derivatives $\dot{\mathbf{x}}^\mathrm{obj}(t), \ddot{\mathbf{x}}^\mathrm{obj}(t), ...$ .
For the sake of simplicity, we omit referring explicitly to the time derivatives in the future.
The notation $\mathbf{x}^\mathrm{obj}$ refers to the fully described trajectory $\mathbf{x}^\mathrm{obj}(t), \dot{\mathbf{x}}^\mathrm{obj}(t), \ddot{\mathbf{x}}^\mathrm{obj}(t), ...$ :
\begin{equation}\label{eqn:x_function_s_and_a}
	\mathbf{x}^\mathrm{obj} = f\left(\mathbf{s}^\mathrm{obj}_{t_0}, a^\mathrm{obj}_{t_0}, ..., a^\mathrm{obj}_{t_0+\Delta t_\mathrm{plan}}\right) 
\end{equation}
A trajectory ensemble consists of one trajectory per traffic participant $\mathbf{X} = (\mathbf{x}^\mathrm{ego}, \mathbf{x}^\mathrm{obj1}, ...)$.
Consequently, a sequence of action vectors $\mathbf{a}$ from initial states $\mathbf{s}$ can be described as such a trajectory ensemble $\mathbf{X}$.

In contrast to the classical MDP formulation, our deterministic MMDP approach is not restricted to stationary policies of other agents.
Further, it has the advantage that the action set of every traffic participant can be easily restricted or even discretized, not only the one of the ego vehicle.
Instead of focusing on an estimate of a probability density function $P_\mathbf{a}(\mathbf{s}, \mathbf{s'})$, we now focus on estimating the future actions of other traffic participants.
Our solution to this MMDP is presented in section \ref{sec:optimization_problem}
.

\section{Related Work}

In this section, we review the related work concerning motion planning for mixed traffic in general, uncertainty and safety guarantees.

\subsection{Motion Planning for Mixed Traffic}

First, we review related approaches for mixed traffic scenarios from the previously introduced MDP perspective.

\paragraph{Independent Prediction}
Many motion planning approaches solve this optimization problem using the following assumption:
Other traffic participants perform actions $a^\mathrm{obj}$ that can be predicted independently.
Consequently, given an object's initial state, all future states, respectively its whole future trajectory $\mathbf{x}^\mathrm{obj}$ can be calculated by an upstream prediction module.
In other words, they assume policies for other agents that only depend on the object's own state and the environment $a^\mathrm{obj} = \pi^\mathrm{obj}(\mathbf{s}^\mathrm{obj}, \mathbf{s}^\mathrm{env})$.
This assumption, that the ego state has no influence on the actions and thus, future states of other traffic participants, largely simplifies the optimization problem.
The expected costs no longer depend on all states and all actions, but only on the ego state $\mathbf{s}^\mathrm{ego}$ and the sequence of ego actions $a^\mathrm{ego}$.
Thus, the expected costs only depend on the ego trajectory $\mathbf{x}^\mathrm{ego}$ (cf. eq. (\ref{eqn:x_function_s_and_a})).
The costs of a planned trajectory, often denoted as the integral over a cost function $L$, can be expressed as 
\begin{equation}
	\int_{t_0}^{t_0 + \Delta t_\mathrm{plan}}
	L(\mathbf{x}^\mathrm{ego}(t))
	\mathrm{d}t ,
\end{equation}
which corresponds to eq. (\ref{eqn:fin_hor_cost}) 
with $\mathbf{s}_{t_i} = (\mathbf{s}_{t_i}^\mathrm{ego}, \mathbf{s}_{t_i}^\mathrm{other}, \mathbf{s}_{t_i}^\mathrm{env})$ where $\mathbf{s}^\mathrm{other}_{t_i}$ and $\mathbf{s}^\mathrm{env}_{t_i}$ are assumed to be known for all $i$ and impose constraints on $\mathbf{x}^\mathrm{ego}(t)$. $\mathbf{s}^\mathrm{ego}_{t_i}$ is given by $\mathbf{s}^\mathrm{ego}_{t_0}$ and $a_{t_i} = a^\mathrm{ego}_{t_i} \; \forall i$.
The optimal solution $\mathbf{x}^\mathrm{ego, *}$ thus can be found via the minimization of this integral: 
\begin{equation}
	\mathbf{x}^\mathrm{ego, *} = \argmin_{\mathbf{x}^\mathrm{ego}} \int_{t_0}^{t_0 + \Delta t_\mathrm{plan}} L(\mathbf{x}^\mathrm{ego}(t)) \mathrm{d}t .
\end{equation}
This approach has for example been applied by Ziegler et al. and has proven to work for a large variety of scenarios \cite{ZieglerAl2014ITSMag}, \cite{ZieglerAl2014TrajPlanning}.
However, as described by the authors, the trajectories were rather over-conservative and sometimes close to the one of human learners starting with driving lessons.
This behavior can be explained mainly by the impact of the 
unknown evolution of the scene $(\mathbf{s}^\mathrm{other})_{t>t_0}$ that depends on the own action $a^\mathrm{ego}$, respectively $\mathbf{x}^\mathrm{ego}$.
Further, the uncertain perception of the status quo $\mathbf{s}_{t_0}$ was dealt with by using security margins. 

\smallskip
\paragraph{Single Agent MDP}
Shalev-Shwartz et al. \cite{mobileye_whitepaper_2017} explicitly model the problem as a single agent MDP.
For the sake of comparability, we use a cost-based instead of a reward-based formulation.
They define the estimated \textit{finite horizon} costs (cf. eq. (\ref{eqn:fin_hor_cost})) called $Q$, regarding the ego action $a = a^\mathrm{ego}$:
\begin{equation}
	Q(\mathbf{s},a) = \min_{(a_{t_1}, ..., a_{t_0 + \Delta t_\mathrm{plan}})} 
	\sum_{i = 0}^{N_{\mathrm{a, pl}}} 
		g(\mathbf{s}_{t_i}, \mathbf{s}_{t_{i+1}}, a_{t_i})
\end{equation}
\begin{equation*}
	\mathrm{s.t.} \; \mathbf{s}_{t_0}=\mathbf{s}, a_{t_0} = a, \forall t_i, \mathbf{s}_{t_{i+1}} = f(\mathbf{s}_{t_i}, a_{t_i})
\end{equation*}
assuming a known, deterministic function $f$ for the state transition.

The first approach would be to seek for the optimal policy $\pi(\mathbf{s})$ via minimizing $Q$:
\begin{equation}
	\pi(\mathbf{s}) = \argmin_a Q(\mathbf{s}, a)
\end{equation}
But since the state space quickly explodes using geometric trajectories, they introduce semantic actions and combine them to so-called \textit{options} or \textit{meta-actions}.
The quality of performing an option is approximated by constructing geometrical trajectories $(\mathbf{s}_{t_0}, a_{t_0}),...,(\mathbf{s}_{t_{N_{\mathrm{a, pl}}}}, a_{t_{N_{\mathrm{a, pl}}}})$ and calculating the \textit{finite horizon} costs

\begin{equation}
	\frac{1}{\Delta t_\mathrm{plan}} \sum_{i = 0}^{N_{\mathrm{a, pl}}} 
		g(\mathbf{s}_{t_i}, \mathbf{s}_{t_{i+1}}, a_{t_i})
\end{equation}
for every option, normalized by the planning horizon $\Delta t_\mathrm{plan}$.

This approach, together with the uncertainty considerations from the following subsection, resolves the challenge of \textit{uncertain perception}.
The authors propose to solve the \textit{uncertain future} challenge by replanning with a high frequency in order to cancel out modeling errors in the dynamics of other agents.
While this might suffice for minor errors in dynamics modeling, different decisions of other agents, such as "who goes first when desired paths cross" cause completely different future states. 
The latter is not addressed by the authors.

Zhan et al. \cite{zhan_NCDS_2016} also model the problem as a single agent MDP.
They do not assume deterministic state transitions regarding the state of other traffic participants.
Instead, they focus on binary decisions on complementary events.
These result from two maneuver options of the other vehicle, for example whether to yield or not at an intersection.
The probabilities are computed using a logistic regression model.
For every option, the future trajectory of the ego vehicle is computed using the approach of Ziegler et al. \cite{ZieglerAl2014TrajPlanning}.
The motion plan for either option has to be equal for a certain (short) time horizon as long as the probability does not equal zero for one of the options.
This corresponds to postponing the decision until one option has probability zero.
Given the probabilities, the expected finite horizon costs are then calculated using the mean of the costs for the two plans, weighted with their probability.
The optimal plan is found via minimizing the expected finite horizon costs.
While this approach is safe, assuming that probability zero equals physical infeasibility of the option, it does not account for the fact that the ego motion affects the motion of other traffic participants.

\smallskip
\paragraph{Multi Agent MDP}
The approaches of Schulz et al.~\cite{schulz_hirsenkorn_2017} and Hubmann et al.~\cite{hubmann_pomdp_2018}  model the problem as a multi agent MDP.
Both works motivate their approach by referring to the interdependence of the ego motion and the motion of other traffic participants.

The work of Schulz et al. \cite{schulz_hirsenkorn_2017} focuses on the estimation of collective maneuvers, using the concept of trajectory homotopy to semantically distinguish them.
After eliminating infeasible and unlikely behavior, they calculate optimal solutions for every maneuver, using a mixed-integer quadratic program (MIQP) to minimize a quadratic global cost function.
As they focus on maneuver estimation, safety considerations are not included in their work.
Also, the uncertainties in the sensed states are not considered in the cost calculation.
Further, the quadratic cost function that is required by the MIQP solver is a strong restriction regarding the goal of modeling human behavior.

Hubmann et al. \cite{hubmann_pomdp_2018} model the problem as a partially observable MDP (POMDP).
They regard the unknown goal destinations of other traffic participants as hidden variables.
The state transition model is deterministic for the ego vehicle and the other vehicles, while the action is determined using a fixed motion and interaction model for other traffic participants.
For the ego vehicle, the action is determined solving the POMDP using a particle-based solver and Monte Carlo simulation.
With this approach, the ego vehicle can implicitly decide to postpone a semantic decision and anticipate the behavior of other agents, as long as they follow the fixed motion and interaction model.
Due to the high computational cost of solving POMDPs, however, the possible action space of the ego vehicle is very limited to keep it online-capable.
Further, safety guarantees are not included in the approach.

\subsection{Uncertainty}
\label{sec:uncertainty}

Concerning the sensing uncertainty, we shortly review the use of Valiants \textit{probably approximately correct (PAC)} terminology from Shalev-Shwartz et al. \cite{mobileye_whitepaper_2017}:
As the exact state $\mathbf{s}$ cannot be expected to be sensed, they define a sensing function that returns an approximate sensing state $\hat{\mathbf{s}}$ from raw sensor data.
The quality of this sensing state $\hat{\mathbf{s}}$ is assessed regarding its implication on the ego action:
In short, $\hat{\mathbf{s}}$ is said to be probably (w.p. of at least $1 - \delta$) $\epsilon$-accurate, if 
$\Pr \left( Q(\mathbf{s}, \pi(\hat{\mathbf{s}}(x))) \geq Q(\mathbf{s}, \pi(\mathbf{s})) - \epsilon \right) \geq 1 - \delta$ with quality function $Q: \mathcal{S}\times\mathcal{A} \rightarrow \mathbb{R}$ and policy $\pi : \mathcal{S}\rightarrow\mathcal{A}$.

However, there is also an uncertainty in the evolution of the scene.
While the latter is not necessarily directly safety relevant, it is largely affecting driving comfort.
Simply consider a narrowing with two vehicles approaching.
This scenario might easily lead to an unintended deadlock, if the intention of the other traffic participant is neglected.
In existing approaches, this uncertainty is at most modeled in some transition probability.

\subsection{Safety Guarantees}
\label{sec:safety_guarantees}

We agree with Shalev-Shwartz et al., stating that in public road traffic, a single agent cannot ensure absolute safety only by its own behavior \cite{mobileye_whitepaper_2017}.
The avoidance of all situations in which others could cause a collision with a self-driving car is neither meaningful nor possible.
For this reason, Shalev-Shwartz et al. introduce a concept called \textit{Responsibility-Sensitive Safety (RSS)}.
In short, this concept promises, that self-driving cars will never \textit{cause} an accident.
This approach is also followed by Althoff et al. \cite{althoff_reachability_2014}, \cite{Koschi2017SPOT}:
They calculate reachable areas for other traffic participants, yet excluding behavior that would lead to the sole responsibility of the other traffic participant in case of an accident.

The RSS concept promises to ensure safety in a meaningful way, while allowing normal flow of traffic, i.e. not being overcautious. 
By deploying it, one is able to guarantee that a vehicle will not cause a collision within the next one or two seconds \cite{mobileye_whitepaper_2017}, \cite{althoff_reachability_2014}, \cite{Koschi2017SPOT}.

\section{Probabilistic Global Optimum Approach}
\label{sec:optimization_problem}

The main focus of this paper is to facilitate the generation of comfortable, safe and comprehensible trajectories.
Using the MMDP model, we propose the following perspective:
The goal of the ego agent is to minimize the expectation $\mathbf{E}$ of its costs for the complete problem via taking an action at time $t_0$: 
\begin{equation}\label{eqn:actual_problem}
	\argmin_{a^\mathrm{ego}_{t_0}} \mathbf{E}\left(
		\sum_{i=0}^{\infty} g^\mathrm{ego}(\mathbf{s}_{t_i}, \mathbf{s}_{t_{i+1}}, \mathbf{a}_{t_i})
	\right) .
\end{equation}

With this model, it is obvious that the optimal action $a^\mathrm{ego}_{t_0}$ cannot be found without imposing further assumptions or simplifications.
The problem even might be ill-posed, if the goal cannot be reached within a finite time $\Delta T$ such that 
$g^\mathrm{ego}(\mathbf{s}_{t_i}=\mathbf{s}_\mathrm{goal}, \mathbf{s}_{t_{i+1}}=\mathbf{s}_{t_i}, \mathbf{a}_{t_i}=\mathbf{a}_\mathrm{idle}) =0 \; \forall t_i > t_0 + \Delta T$. 
Thus, instead of strictly optimizing or applying machine learning to a very simplified problem, we propose to strive for an approximate solution of a problem that is closer to the desire of passengers.
Having said that, safety must of course be guaranteed.
Its separate treatment using analytical methods can be motivated by the very high cost that collisions impose.

In the following, we shortly describe the basic idea behind the approach, before explaining our way of dealing with the uncertain perception and the uncertain prediction along with giving safety guarantees.

\subsection{Basic Idea}

Firstly, the horizon of the planning problem is largely reduced by using a state of the art navigation approach to compute the approximate path to be traveled, similar to a human using a navigation device.

Within this narrowed planning horizon, as motivated previously, our main goal is to plan safe trajectories.
Obviously, we will never put safety at risk in order to gain comfort.
Following the goal of minimizing our expected costs (cf. eq. (\ref{eqn:actual_problem})), 
our approach is to drive mainly comfortable while risking rare uncomfortable maneuvers.
The latter might for example be a response to very unlikely behavior of other traffic participants.
Safety is always guaranteed by following the RSS concept (cf. Section \ref{sec:safety_guarantees}), assuming a PAC sensing system (cf. Section \ref{sec:uncertainty}).

The basic idea behind our driving policy is to behave comprehensible, i.e. human-like, and thus allow for cooperation with other traffic participants.
To achieve this, we use a cost functional accounting for comfortable dynamics and also perceived safety (cf. \cite{naumann2017cooperativePlanning}).
Further, we explicitly model the strategy of other traffic participants, and determine an estimate for the model compliance of each participant.
When having a high model compliance, we are able to anticipate the actions of others and thus, the evolution of the scene, with less uncertainty.
Only when the model compliance is too low, we change to a conservative driving strategy.
In other words, we have less discrepancy between the estimated future states $\hat{\mathbf{s}}_{t_i}$ and the actual future states $\mathbf{s}_{t_i}$ in most situations.
As in game theory (e.g. prisoner's dilemma), optimizing jointly can unveil solutions that are globally optimal if the agents trust each other.
The latter can be achieved by behaving presumably.
Hence, by and large, these solutions are better for all traffic participants.

\subsection{Ideal Case: Global Optimum Unique and Pursued}

In order to model the strategy of others, we assume that they also seek for a comfortable and safe trajectory. 
Consequently, we investigate their costs for potential future states.
Even though we can only directly control our ego vehicle, 
we seek to find a trajectory ensemble close to the global optimum. 

As presented in our previous work \cite{naumann2017cooperativePlanning}, the cost functional for this globally optimal ($\mathrm{go}$) finite horizon approach is:
 \begin{equation}
	G_{\mathrm{go}}(\mathbf{X})
	=
	\sum_{i} \left( G_{i, 0}(\mathbf{x}^i) + \sum_{j \neq i} G_{i, j}(\mathbf{x}^i, \mathbf{x}^j) \right)
\end{equation}
with singleton trajectory costs $G_{i, 0}(\mathbf{x}^i)$ for vehicle~$i$
and pairwise trajectory costs $G_{i, j}(\mathbf{x}^i, \mathbf{x}^j)$ for vehicle~$i$ due to vehicle~$j$.

The optimal trajectory ensemble $\mathbf{X}$ can then be found via minimizing $G_{\mathrm{go}}(\mathbf{X})$.
As stated previously, we propose to strive for an approximate solution of this highly non-convex problem and therefore apply sampling.
The best solution out of $K$ samples $\{\mathbf{X}^1, ..., \mathbf{X}^K\}$ simply is $\mathbf{X}^{k^{*}}$ such that
\begin{equation*}
	 G_{\mathrm{go}}(\mathbf{X}^{k^{*}}) < G_{\mathrm{go}}(\mathbf{X}^{k}) \; \; \forall k \in [1,K]\setminus k^{*}.
\end{equation*}

\subsection{Model Compliance Consideration}
\label{sec:model_compliance_consideration}

In this approach, we are also faced with uncertainties, regarding the accuracy of our model:
The sensed state $\mathbf{s}_{t_0}$, building the start of the future trajectories $\mathbf{x}$ (cf. eq. (\ref{eqn:x_function_s_and_a})), is subject to uncertainties.
Further, the action sequence $\mathbf{a}_{t_i}, i > 0$ of all traffic participants but the ego vehicle is only an expectation.
If a traffic participant deviates from our globally optimal plan, obvious and possible explanations are 
(a) he does not act according to a (stationary) policy $\pi(s)$,
(b) we estimated his costs or destination wrongly or
(c) he estimated our costs wrongly.
For the sake of simplicity, the desired paths or destinations of the traffic participants are assumed to be known for now. 
This assumption is relaxed later on.

Hence, the optimal costs are only reached with a certain probability, denoted $p(\mathrm{go})$, while violations of the previous assumptions can cause higher costs.
The refined cost expectation with model compliance consideration ($\mathrm{mcc}$) is 
\begin{equation}\label{eqn:expected_cost_with_probability}
 	\mathbf{E}\left( G_{\mathrm{mcc}} \right)
 	=
	p(\mathrm{go}) G_{\mathrm{go}}
	+
	(1-p(\mathrm{go}))\mathbf{E}\left( G_{\neg\mathrm{go}} \right) ,
\end{equation}

containing two unknowns: the probability $p(\mathrm{go})$ that the globally optimal plan is followed and the expected costs $\mathbf{E}( G_{\neg\mathrm{go}})$ in case the latter is not followed.

As neither humans nor this approach strive for strict optimality, we approximate equation (\ref{eqn:expected_cost_with_probability}) as follows: the probability $p(\mathrm{go})$ is not the probability 
that the globally optimal plan $\mathrm{go}$ is exactly followed.
Instead, it is the probability that our model is compliant $p(\mathrm{go})=p(\mathrm{mc})$ and thus any plan that is \textit{close to} $\mathrm{go}$ is pursued.
For the second unknown, the expected costs in case of the non optimal plan $\mathbf{E}( G_{\neg\mathrm{go}})$, we consider a conservative fallback plan ($\mathrm{cfb}$), such as refraining from entering a potential collision zone. 
We consider these costs as $G_{\neg\mathrm{go}}=G_{\mathrm{cfb}}$.
These considerations yield
\begin{equation}\label{eqn:expected_cost_with_probability_new}
 	\mathbf{E}\left( G_{\mathrm{mcc}} \right)
 	=
	p(\mathrm{mc}) G_{\mathrm{go}}
	+
	p(\neg\mathrm{mc})\mathbf{E}\left( G_{\mathrm{cfb}} \right) .
\end{equation}

In order to check for model compliance, we first check for ambiguity, i.e. how much uncertainties in perception and prediction affect the global optimum.
Therefore, we consider the following marginal cases:
Every traffic participant $i$ can be a dynamic driver or a defensive driver.
That is, the cost parametrization is chosen accordingly.
We check whether all permutations of the cost parametrizations yield the optimum in the same homotopy class, employing  the concept of trajectory homotopy of \cite{schulz_hirsenkorn_2017} and \cite{Bender2015combinatorial}.
An example for different homotopy classes can be seen in Figure \ref{fig:page1fig_b} and for equal homotopy classes in \ref{fig:page1fig_c}.
Further, the marginal cases of the PAC sensing system are included here in order to consider the marginal case of the worst combination of uncertain perception and uncertain prediction.

If the marginal cases yield optima in different homotopy classes, the optimal solution of the situation is ambiguous and thus, the probability that our model describes the evolution of the situation correctly $p(\mathrm{mc})$ is lowered.
Instead of trying to force the solution in our preferred homotopy class, while assuming that others will certainly try to avoid a collision and thus follow our desire, we consider a more conservative fallback plan.

If the costs of the fallback plan $\mathrm{cfb}$ are very low, for example because the potential collision zone is far away, we pursue $k^{*}$ and do not act overcautiously.
In this case, it is still possible that the situation becomes unambigous and our model describes the evolution of the situation correctly.
If, however, the costs of $\mathrm{cfb}$ are high, while the probability $p(\neg\mathrm{mc})$ is also high, we do no longer trust in the model compliance.
Hence, we change to the conservative fallback plan.
When the homotopy class changes, we try to find the globally optimal solution again and build trust in this solution.
Eventually, we will build trust in the new homotopy class, so that we switch back to $\mathrm{go}$ again, or the situation will be solved conservatively, i.e. with less risk, e.g. because we drive slowly and $\mathbf{E}( G_{\mathrm{cfb}})$ is low.

Given an initial globally optimal plan with one distinct homotopy, $p(\mathrm{mc})$ is set to a high value.
From then on, we constantly check on every state update, whether the other traffic participants' behavior lies within the expected range. 
Note, that this is a violation of the Markovian assumption.
However, one could introduce additional state variables describing the trust or the non-compliance of an agent with our assumption.
If we detect a violation of this explicable behavior, this also increases the uncertainty in our estimation of others' driving strategy and thus the estimates for their future states.
This behavior might for example be due to a maneuver that was not anticipated, such as parking along the road or avoiding an unforeseen obstacle, or due to a very dynamic driving policy that exceeds the scope of our considerations.
It can be detected through a rise in the costs of the globally optimal plan, through a violation of the state limits determined by the marginal cases, or through a shift of the time at which traffic participants enter and leave a potential collision zone.
In case the deviation questions our homotopy choice, i.e. an agent that is supposed to go second accelerates, we decrease $p(\mathrm{mc})$.

\subsection{Intention Consideration at Intersections}
\label{sec:intention_consideration}

At intersections, agents mostly have the choice between several routes.
Neglecting emergency situations, this decision can be considered independent of the behavior of other agents, as it is made in an upstream navigation layer, also for human drivers \cite{donges1999conceptual}.
Thus, this route estimation can also be made by an upstream module, as presented by Petrich et al. \cite{petrich_prediction_2014}.
As input to the planner, we receive different route combinations $\mathcal{R}=\{r_1, r_2, ...\}$ for the ensemble of traffic participants along with their probabilities $\{p(r_1), p(r_2), ...\}$.
With this input, we can relax the assumption of knowing the desired destination of the traffic participants, considering the cost expectation for multiple routes with model compliance check ($\mathrm{mcc,mr}$): 
\begin{equation}\label{eqn:total_cost_multiple_paths}
	\mathbf{E}\left( G_{\mathrm{mcc,mr}} \right)
	=
	\sum_{r \in \mathcal{R}}  p(r) \mathbf{E}\left(G_{\mathrm{mcc, route\;} r}(\mathbf{X}^{r})\right) 
\end{equation}
with $\mathbf{E}\left( G_{\mathrm{mcc, route\;} r} \right)$ from eq. (\ref{eqn:expected_cost_with_probability}) and $\sum_{r \in \mathcal{R}}  p(r) = 1$.
Note, that for unique paths, this equation can be simplified to eq. (\ref{eqn:expected_cost_with_probability}).
Note further, that the ego trajectory $\mathbf{x}^{r, i=\mathrm{ego}}$ must be identical for all $r \in \mathcal{R}$ up to the output time of the subsequent planning step.
With this condition, analog to Hubmann et a. \cite{hubmann_pomdp_2018}, we are able to postpone decisions as long as multiple route hypothesis are evident.
Similar to human drivers, instead of over-cautiously reacting to all possible predictions, we act accordingly for the likely predictions, and perform rather sharp maneuvers, when we encounter an unlikely action.

\subsection{Safety Consideration}
As mentioned in our previous work, approaches that predict cooperative behavior of other traffic participants potentially decrease safety, as the prediction might be wrong. 
This risk is particularly high when predicting that others yield.
Obviously, a cooperative approach should not lead to a more risky driving policy.
On the other hand, absolute safety cannot be guaranteed in road traffic in any case.
Thus, we follow the notion of the previously introduced RSS \cite{mobileye_whitepaper_2017} and guarantee to not cause accidents.

As the previously introduced approach of solving a complex MMDP cannot give guarantees regarding safety or even finding a feasible solution, the approach needs to be backed up.
Thus, we apply the RSS concept of analytically computing safety margins regarding the physically feasible and lawful motion of all other agents parallel to the cooperative planner.
We calculate these safety margins to other traffic participants and potentially occluded areas, taking our own computation time and others' reaction time into account, as described by \cite{mobileye_whitepaper_2017}.
As soon as we observe a violation of those safety margins, we react with the \textit{appropriate response} (cf. \cite{mobileye_whitepaper_2017}) as an emergency plan, which in most cases is a full deceleration, until the safety margin is satisfied again.
This computationally cheap check can be done with a very high frequency, allowing to use a small ego reaction time.
The latter results in a smaller safety distance and thus facilitates a larger scope for cooperative maneuvers. 
\section{Results and Evaluation}
\label{sec:results_and_evaluation}

\begin{figure}
	\includegraphics[trim={8cm 0.8cm 7cm 0},clip,width=\linewidth]{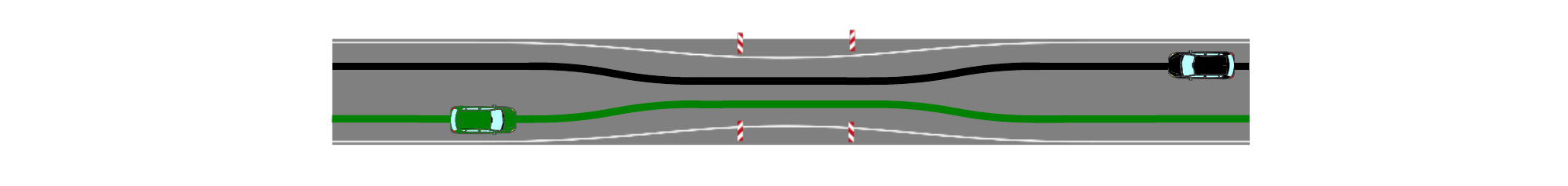}
	\caption{Narrowing, without signposted right of way and with one predefined path per agent.}
	\label{fig:scenario_narrowing}
\end{figure}

\begin{figure}
	\centering
	\includegraphics[trim={6.5cm 0 6.5cm 0},width=0.6\linewidth]{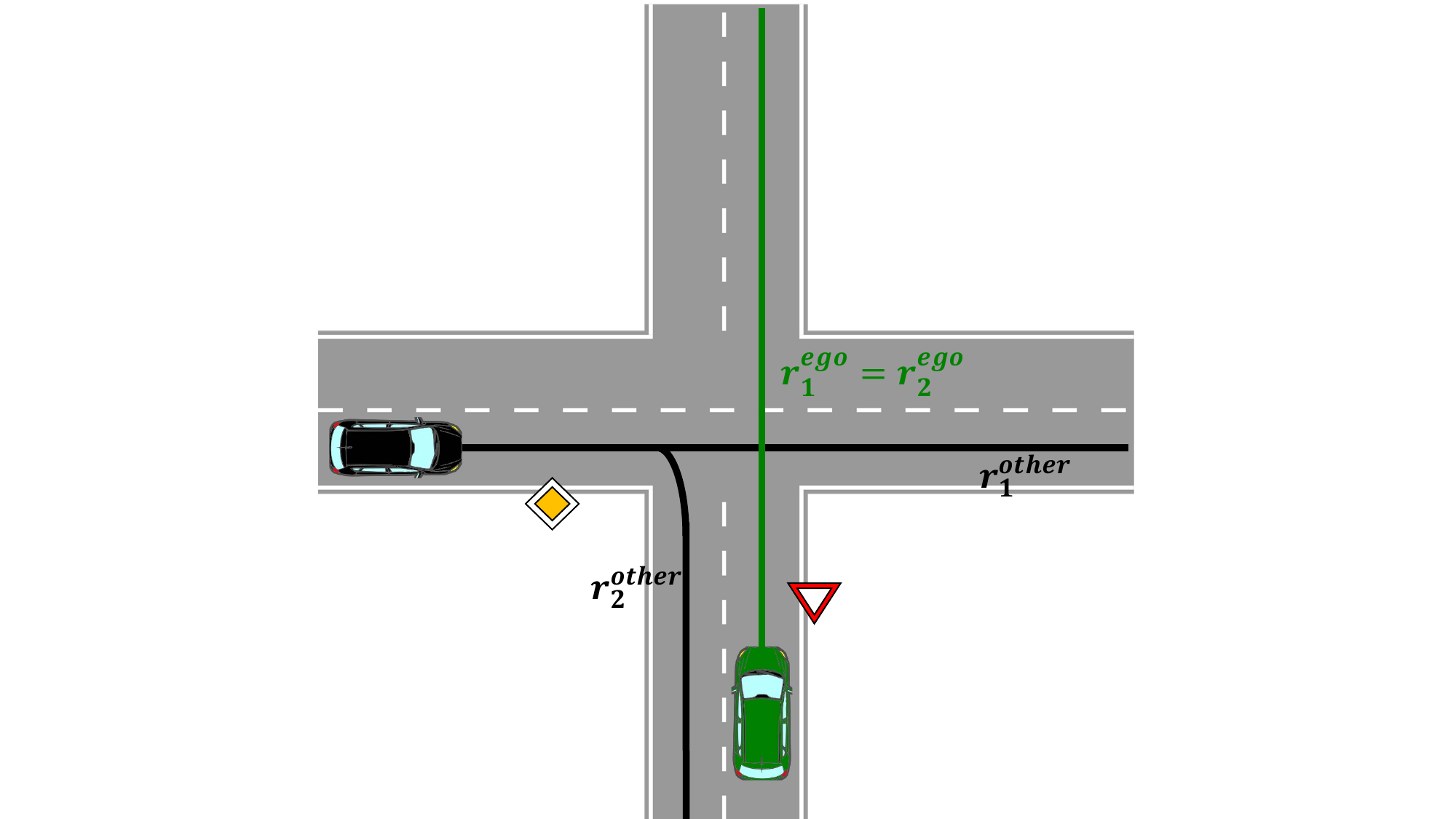} 
	\caption{Intersection: The other traffic participant (black) has the right of way and can drive straight on path $r_1^{\mathrm{other}}$ or turn right on path $r_2^{\mathrm{other}}$. The path of the ego vehicle (green) is known.}
	\label{fig:scenario_intersection}
\end{figure}

The approach is evaluated for two scenarios, passing an intersection and passing through a narrowing of the road (cf. Figures \ref{fig:scenario_narrowing} and \ref{fig:scenario_intersection}).
The narrowing shows the benefits of the approach regarding scenarios where the right of way is not predefined.
At the intersection, the benefits of the intention consideration are shown.

\begin{figure}
	\includegraphics[trim=0cm 1cm 5cm 3cm,clip, scale=0.41]{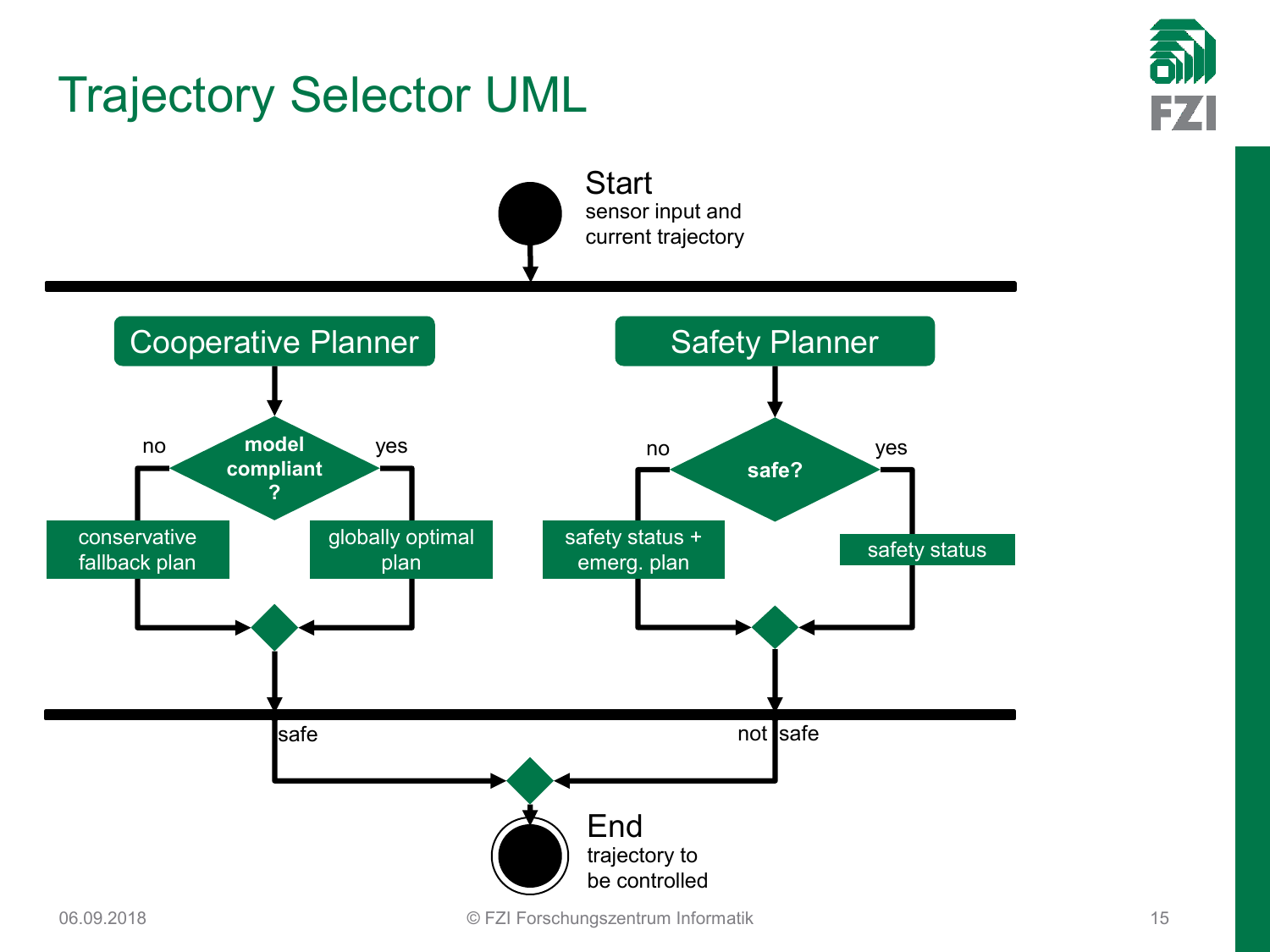}
	\caption{Simplified overview of the python implementation of the planner.}
	\label{fig:ablaufdiagramm}
\end{figure}

In order to evaluate our approach, we implemented it in our ROS based simulation framework \textit{CoInCar-Sim} \cite{Naumann2018Sim} using python.
A simplified overview is given in Fig. \ref{fig:ablaufdiagramm}.
Many cooperative scenarios, including the previously mentioned, have highly constrained driving corridors.
Thus, we applied path-velocity-decomposition, as introduced by \cite{kant1986PVD}.
The paths were generated from a polygon depicting the centerline of the lane. 
Further, we used a naive jerk sampling approach to generate multiple trajectory candidates per traffic participant.
In order to do so, we discretized the path in space ($0.5m$) and determined potential collision zones.
Jerk samples that violate the kinematic restrictions are replaced by the respective marginal jerk, thus implying a bias on marginal cases.
We assumed a planning deadtime of $1s$ and chose $\Delta t_\mathrm{action}=1s$ accordingly.
Thus, an action planned at time $t_i$ assumes its motion in $[t_i, t_{i+1})$ to be known and is effective in the interval $[t_{i+1}, t_{i+2})$.

We chose the planning horizon dynamically, such that the potential collision zone is passed by both traffic participants, but bounded by $\Delta t_\mathrm{plan,max}=35s$.
For the cost computation, we consider the velocity, the normal and the tangential acceleration.
For the perceived safety, we consider the time of zone clearance, depicting the time that elapses between the first vehicle leaving potential
collision zone and the second vehicle entering this area.
A more detailed overview is given in our previous work \cite{naumann2017cooperativePlanning}.
Instead of introducing \textit{end costs} for unsafe final states $s_{t_N}$ with $N=N_{\mathrm{a, pl}}$, we discarded those samples.
In order to prove safety according to the RSS concept, we compute the safety margins analytically with a frequency of $100\mathrm{Hz}$.
We consider the following cases: 
If we can stop before entering the potential collision zone, our state is considered safe.
If we cannot stop before entering the potential collision zone, but the safe longitudinal distance (Def. 2 of \cite{mobileye_whitepaper_2017}) is  satisfied, our state is considered safe.
If we leave the collision zone according to our pursued plan before the other vehicle is able to enter it, considering its physical limits, our state is also considered safe. %

The costs expectation $\mathbf{E}\left( G \right)$ is calculated from eq. (\ref{eqn:total_cost_multiple_paths}) with $G_{\mathrm{go}}$ from \cite{naumann2017cooperativePlanning}.
The probability $p(\mathrm{mc})$ is initialized to be $50\%$. %
At the beginning of each subsequent planning, it is investigated whether the other traffic participant behaved within the expected boundaries.
This is done via checking whether the total costs do not raise by more than factor $1.5$.
If they stay below, $p(\neg\mathrm{mc})$ is lowered by $50\%$, if not, $p(\mathrm{mc})$ is lowered by $50\%$.
If the conservative maneuver after the subsequent planning step would impose high costs\footnote{In our case, costs of a deceleration with more than $3\frac{m}{s^2}$.} $\mathbf{E}\left( G_{\mathrm{cfb}} \right)$, we only continue the plan $\mathrm{go}$ if its probability of model compliance is high: $p(\mathrm{mc})>95\%$.
Otherwise, we change to the conservative maneuver.

\begin{figure}
	\includegraphics[width=\linewidth]{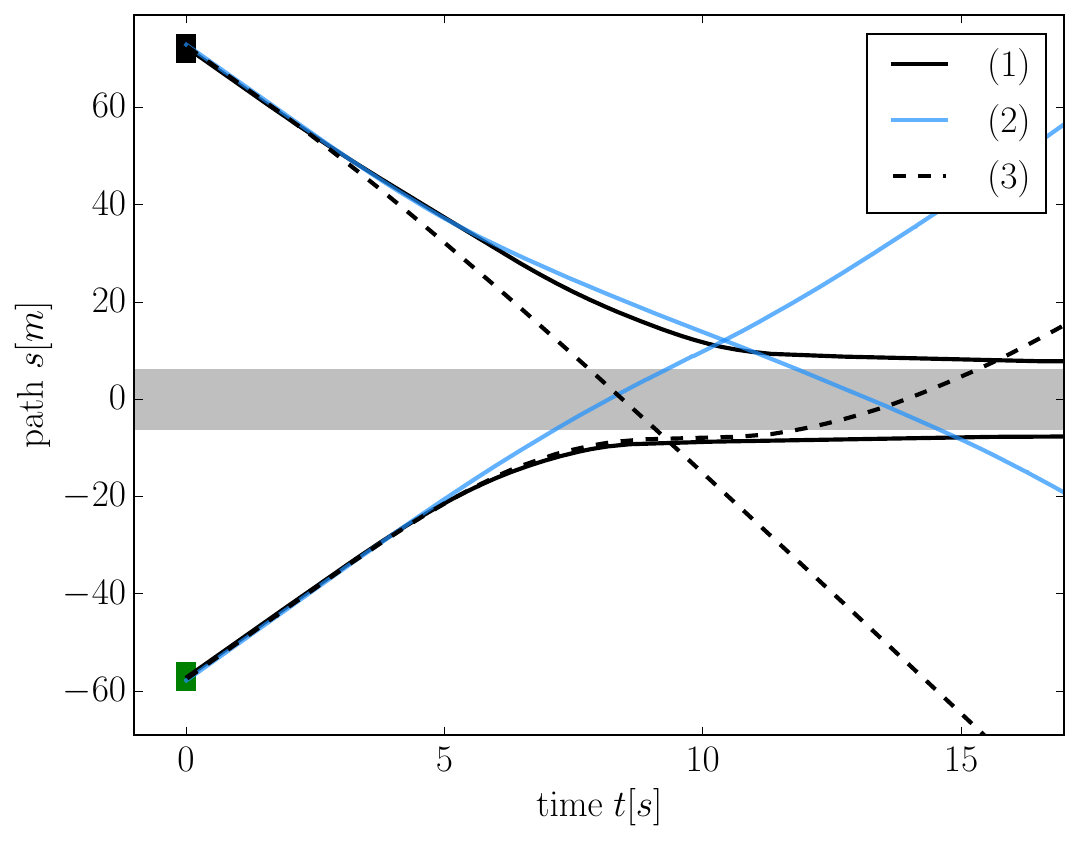} 
	\caption{Solutions to the narrowing scenario (Fig. \ref{fig:scenario_narrowing}): Classical approaches that mutually predict the other traffic participant with constant velocity result in a deadlock (1) by stopping in front of the potential collision zone (grey). With the proposed approach the vehicles mutually include their behavior and thus act globally optimal (2). In (3) the right vehicle is driving ignorantly. The left vehicle, running our approach, detects this and reacts early by yielding.}
	\label{fig:narrowing_all_history}
\end{figure}

\subsection{Narrowing}

At the narrowing, we consider three cases.
The driven trajectories are visualized in Figure \ref{fig:narrowing_all_history}:
$(1)$ Two automated vehicles, using a classical approach with a constant velocity prediction, as in \cite{ZieglerAl2014TrajPlanning},
$(2)$ two automated vehicles using our approach, and $(3)$ one vehicle using our approach together with an ignorant driver that does not consider us.
Even though the situation seems obvious to humans, classical approaches lead to a deadlock in this situation, as they neglect their mutual influence on each other. Their optimization problem yields two local optima, none of which is close to the actual global optimum.
With the proposed approach, however, the mutual influence is considered.
Thus, for this unambiguous situation, a global optimum is found and pursued by both vehicles.
As claimed, this behavior is comfortable, also with respect to perceived safety, and comprehensible for other traffic participants.
Further, it is provably safe.
If we detect violations of our assumptions considering the behavior of others, $p(\neg\mathrm{mc})$ increases and we choose a more conservative plan.
Hence, we even react comfortably to an ignorant driver, while other approaches would have to perform sharp maneuvers if their assumptions are violated.

\subsection{Intersection}

\begin{figure}
	\includegraphics[width=\linewidth]{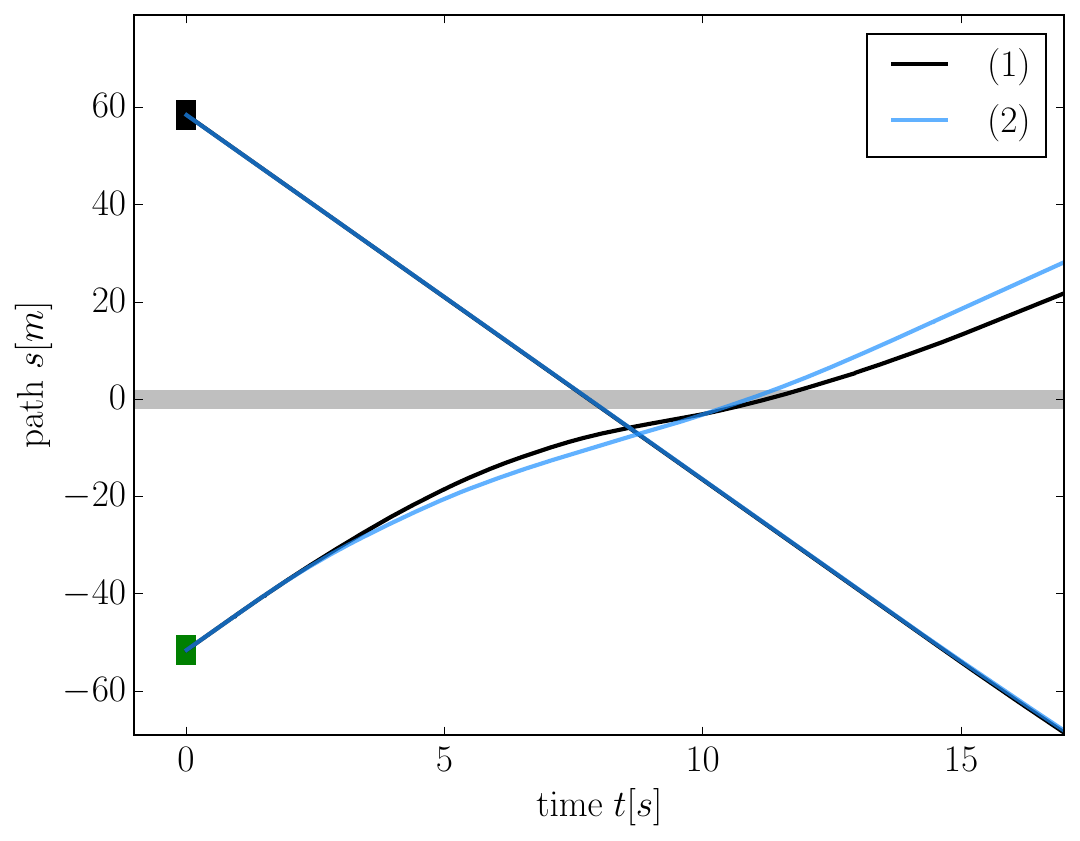} 
	\caption{Solutions to the intersection scenario (Fig. \ref{fig:scenario_intersection}): In both cases, the other vehicle (starting at $s=60\mathrm{m}$) drives straight on $r_1$ (potential collision zone in grey) with constant velocity. In case (1), the planning module was fed the (wrong) information $p(r_2)=99\%$. Thus, it reacts "surprised" and brakes late. In case (2) it was fed  $p(r_1)=99\%$. Thus, it pursues the globally optimal plan for $r_1$ from the beginning.}
	\label{fig:tjunct_all_history}
\end{figure}

At the intersection, the route of the other vehicle is not known but estimated by an upstream module.
The trajectories driven for different probabilities of the two routes are visualized in Figure \ref{fig:tjunct_all_history}.
We are able to reproduce the key result of Zhan et al. \cite{zhan_NCDS_2016} and Hubmann et al.  \cite{hubmann_pomdp_2018}:
We implicitly postpone a decision, while acting in order to minimize the expected costs.
Still, we consider the mutual influence and obey the traffic rules without defining homotopy constraints.

\FloatBarrier

\section{Conclusions and Future Work}
\label{sec:conclusion}

In this paper, we reviewed the problem statement of behavior generation and motion planning for automated vehicles.
We propose to model the problem as an MMDP with deterministic state transitions.
This model allows to incorporate the prediction of other traffic participants in an integrated approach.
Further, the consequences of wrong assumptions concerning the behavior of other traffic participants are explicitly considered.
Also, estimations of upstream modules such as a route prediction for other traffic participants are made use of.
The resulting behavior of a vehicle following the proposed approach is comfortable, safe and comprehensible.
Scenarios that are obvious to humans are solved human-like: E.g. if one vehicle is closer to a narrowing than the other, it drives first.
If we are confident, that a vehicle that has priority will not intersect our path, e.g. by turning right in front of us, we drive as we would not have to give way but still keep this option open, and react more harshly in case our assumption was wrong.
Traffic rules such as the right of way are modeled by regarding the time of zone clearance, instead of explicitly excluding certain homotopy classes.

The authors intend to further pursue the approach:
In order to be able to stick to the cooperative plan as often as possible, 
an important future work is to improve our model of comfort and perceived safety in the cost functional.
For this purpose, simulator studies with human drivers can be conducted or trajectory datasets can be analyzed.
Further, we intend to investigate different directed sampling methods and port the algorithm to our probe vehicle "Bertha" in order to test the approach in real mixed traffic.

\section*{Acknowledgements}
We gratefully acknowledge support of this work by the Tech Center a-drive, 
the Profilregion Mobilitätssysteme Karlsruhe 
and by the Deutsche  Forschungsgemeinschaft
(German  Research  Foundation)  within  the  Priority  Program  
``SPP  1835 Cooperative  Interacting  Automobiles''.

\bibliographystyle{IEEEtran}
\bibliography{sections/references}

\end{document}